%% file: main.tex
\documentclass{article}
\usepackage{microtype}
\usepackage{graphicx}
\usepackage{subfigure}
\usepackage{booktabs}
\usepackage{hyperref}
\usepackage{xspace}
\usepackage{makecell}
\usepackage[small]{caption}
\usepackage{booktabs, multirow}
\usepackage{array}
\usepackage{textcomp}
\usepackage{xcolor}

\usepackage{kotex}
\usepackage{enumitem}
\usepackage[accepted]{icml2020}

\icmltitlerunning{Submission and Formatting Instructions for ICML 2020}

\begin{document}
\twocolumn[
\icmltitle{\fontsize{13}{20}\selectfont XOR Mixup: Privacy-Preserving Data Augmentation for One-Shot Federated Learning}
%\icmlsetsymbol{equal}{*}

\begin{icmlauthorlist}
\icmlauthor{MyungJae Shin}{to}
\icmlauthor{Chihoon Hwang}{goo}
\icmlauthor{Joongheon Kim}{ec}
\icmlauthor{Jihong Park}{ed}
\icmlauthor{Mehdi Bennis}{def}
\icmlauthor{Seong-Lyun Kim}{abc}
\end{icmlauthorlist}

\icmlaffiliation{to}{Seoul National University Hospital, Seoul, Korea}
\icmlaffiliation{goo}{Department of Computer Science and Engineering, Chung-Ang University, Seoul, Korea}
\icmlaffiliation{abc}{School of Electrical \& Electronic Engineering, Yonsei University, Seoul, Korea}
\icmlaffiliation{ed}{School of Information Technology, Deakin University, Geelong, Australia}
\icmlaffiliation{def}{Centre for Wireless Communications, University of Oulu, Finland}
\icmlaffiliation{ec}{School of Electrical Engineering, Korea University, Seoul, Korea}

\icmlcorrespondingauthor{Jihong Park}{jihong.park@deakin.edu.au}
\icmlcorrespondingauthor{Joongheon Kim}{joongheon@korea.ac.kr}
\icmlkeywords{Machine Learning, ICML}

\vskip 0.3in
]
\printAffiliationsAndNotice{\icmlEqualContribution}
\begin{abstract}
User-generated data distributions are often imbalanced across devices and labels, hampering the performance of federated learning (FL). To remedy to this non-independent and identically distributed (non-IID) data problem, in this work we develop a privacy-preserving XOR based mixup data augmentation technique, coined \emph{XorMixup}, and thereby propose a novel one-shot FL framework, termed \emph{XorMixFL}. The core idea is to collect other devices' encoded data samples that are decoded only using each device's own data samples. The decoding provides synthetic-but-realistic samples until inducing an IID dataset, used for model training. Both encoding and decoding procedures follow the bit-wise XOR operations that intentionally distort raw samples, thereby preserving data privacy. Simulation results corroborate that XorMixFL achieves up to $17.6$\% higher accuracy than Vanilla FL under a non-IID MNIST dataset.
\end{abstract}

\input{introduction_JH}
\input{related}
\input{method}

\input{experiment}

\input{conclusion}

\bibliographystyle{icml2020}
\bibliography{reference}

\end{document}

%% file: introduction_JH.tex
\section{Introduction}\label{sec:intro}
% ML -> Problem -> Reason for FL -> FL -> FL problems -> Recent FL -> Solutions 

Securing more data is essential in imbuing more intelligence into machine learning (ML) models. In view of this, the problem of utilizing the sheer amount of user-generated private data has attracted significant attention in both academia and industry \cite{park2018wireless,kairouz2019advances,park2020extreme}. Federated learning (FL) is one promising solution based on exchanging model parameters among devices without sharing raw data, thereby preserving data privacy \cite{McMahan2016,Konecny2016,Yang:FLSurvey,Smith:FLSurvey}. While effective under independent and identically distributed (IID) data distributions, the performance of FL is highly degraded under non-IID user-generated data in practice~\cite{Zhao2018,Oh20:CL}. Indeed, when each device has scarce samples of specific labels, the classification accuracy under MINST and CIFAR-10 datasets is degraded by up to $11$\% and $51$\%, respectively, compared to the IID counterparts~\cite{zhao2018federated}.

On this account, in this article we seek for an FL solution coping with non-IID data distributions. Inspired by the Mixup data augmentation technique (Vanilla Mixup) producing a synthetic sample by linearly superpositioning two raw samples~\cite{Zhang2018}, we first propose an \emph{XOR based mixup data augmentation method (XorMixup)} that is extended to a novel FL framework, termed \emph{XorMixFL}. 

\begin{figure*}
  \centering
  \includegraphics[width=0.7\textwidth]{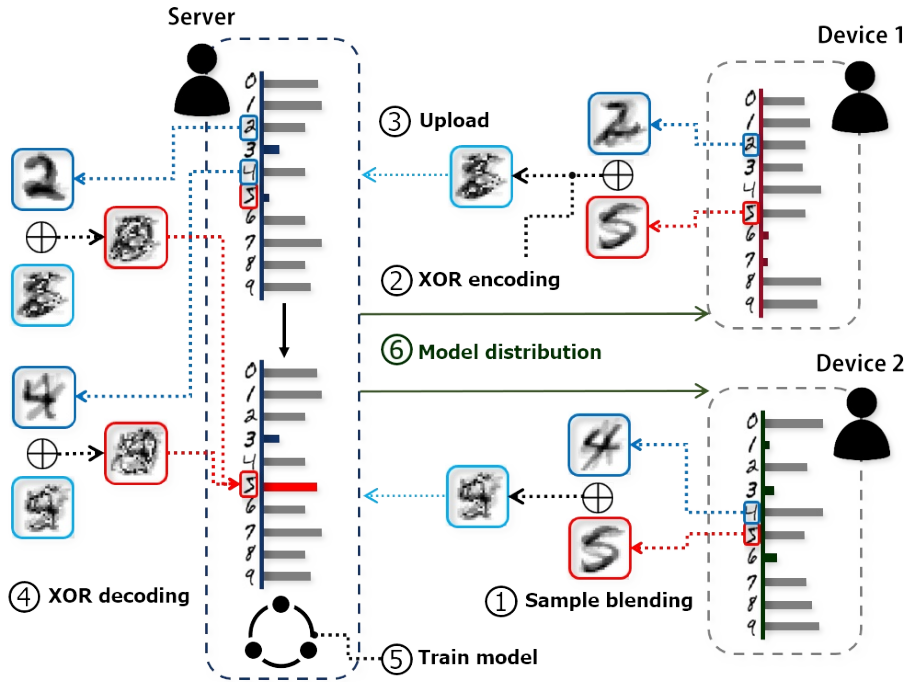}
  \vspace{-3mm}
  \caption{A schematic illustration of \emph{XorMixFL} in which XorMixup data augmentation is used for correcting non-IID data distributions while preserving raw data privacy.}
  \label{fig:overview}
%   \vspace{-5mm}
\end{figure*}

\vspace{-5pt}
\paragraph{XorMixup.}
The key idea is to exploit the flipping property of the bit-wise XOR operation $\oplus$: $(A \oplus B) \oplus B = A$. For two devices $d_1$ and $d_2$, XorMixup is operated as follows.
\vspace{-5pt}
\begin{enumerate}[leftmargin=*]
%   \vspace{-5pt}
  \item[(i)] $d_1$ generates an encoded seed sample $A \oplus B)$ by mixing samples $A$ and $B$ in different labels, transmitted to~$d_2$. 
%   \vspace{-8pt}
  \item[(ii)] $d_2$ decodes $(A \oplus B)$ by mixing $(A \oplus B)$ and its own sample $B'$ having the same label of $B$, via $(A \oplus B) \oplus B'$, producing a synthetic sample $A'$ that is similar to but different from~$A$.  
\end{enumerate}
\vspace{-5pt}
  While both (i) and (ii) preserve raw data privacy between two devices, (ii) improves the synthetic sample's authenticity, increasing one-shot FL accuracy as detailed next.

% % XorMixup: generating a synthetic sample based on XOR operations in the following two steps: 1. (encoding after averaging) each device first linearly superpositions local samples per class, and then combine these averaged samples in different classes using the bit-wise XOR operation; 2. (decoding) the server storing a public dataset combines the encoded sample with its own sample using the bit-wise XOR operation so as to generate a synthetic sample having a single class. The encoding preserves local data privacy at the server's reception, while the decoding improves the authenticity of the generated sample while preserving local data privacy even after decoding.

  \vspace{-5pt}\paragraph{XorMixFL.}
  As illustrated in Fig.~\ref{fig:overview}, by applying XorMixup to a one-shot FL framework having only one communication round~\cite{Guha:19,HybridFL}, each device in XorMixFL uploads its encoded seed samples to a server. The server decodes and augments the seed samples using its own base samples until all the samples are evenly distributed across labels. The server can be treated as one of the devices, or a parameter server storing an imbalanced dataset. Then, utilizing the reconstructed dataset, the server trains a global model that is downloaded by each device until convergence.

\vspace{-5pt}
\paragraph{Contributions.} To the best of the authors’ knowledge, this is the first piece of FL research based on XOR operations for addressing the non-IID data problem. Under a non-IID MNIST dataset, simulation results corroborate that XorMixFL achieves up to $8.13$\% and $17.6$\% higher accuracy than standalone ML and Vanilla FL, respectively. For an ablation study, we additionally propose a baseline one-shot FL (MixFL) whose encoding follows from Vanilla Mixup while ignoring decoding. Compared to MixFL, XorMixFL achieves comparable accuracy while preserving more data privacy, i.e., higher dissimilarity between the augmented and original data samples measured using the multidimensional scaling (MDS) method~\cite{Cox2008}, highlighting the importance of XorMixup in one-shot FL.

%% file: related.tex
\section{Related Work}\label{sec:related}

Existing one-shot FL schemes~\cite{Guha:19,HybridFL} consider that each devices first trains a local model until convergence, and then the server constructs a global model by aggregating the converged local models. This does not take into consideration global data distributions, and is thus vulnerable to the non-IID data problem. By contrast, XorMixFL constructs the global model by training a model with synthetic samples that are uploaded from devices in one communication round while preserving their data privacy.

To preserve privacy while exchanging data samples, homomorphic encryption such as RSA \cite{Craig09:Homomorphic} or differential privacy mechanisms~\cite{Koda:GC20} can be used, with non-negligible computing overhead or accuracy degradation, respectively. Alternatively, XorMixup reduces the accuracy degradation with low complexity by leveraging Mixup data augmentation~\cite{Zhang2018} and XOR operations. The original purpose of Vanilla Mixup is oversampling. XorMixup additionally focuses on its privacy preserving benefit in that the combination distorts raw samples. Furthermore, instead of the linear combination used in Vanilla Mixup, we apply XOR operations that are often used in cipher algorithms for hiding original information \cite{Churchhouse01:book}. 

Several recent works also study FL frameworks based on data sample exchanges~\cite{Jeong18,Jeong:FML19,Oh20:CL}. For one-shot FL under non-IID data, in \cite{Jeong18,Jeong:FML19} a synthetic sample generator is trained after collecting seed samples, which may still violate data privacy. In \cite{Oh20:CL}, seed samples are collected after Vanilla Mixup encoding, for running knowledge distillation operations, rather than one-shot FL. For a comprehensive overview on data sample exchanges in the context of FL, compared to model parameter exchanges \cite{McMahan2016,Konecny2016,KimCL:19,Chen:20019aa,YangQuekPoor:2019aa,Wang:2018aa,Amiri:SPAWC19,elgabli2019gadmm,samarakoon2018federated,chen2018lag} and model output exchanges~\cite{Jeong18,Oh20:CL,Han:Intellisys20,Ahn:PIMRC20,Ahn:ICASSP20}, readers are encouraged to read \cite{Park:2018ab}.

%% file: method.tex
\section{Methodology}\label{sec:method}

In this section, we describes the system model under study and the operations of {XorMixFL}. Consider a one-shot FL system consisting of one server and $\mathcal{U}$ devices, as depicted in Fig.~\ref{fig:overview}. The $i$-th device $u_i$ stores its local dataset $D_i$, where $i \in \{1,2, \dots , \mathcal{U} \}$. Let $c$ indicate the server training and distributing a global model after collecting samples from $\mathcal{U}$ devices in a privacy-preserving way. 

In {XorMixFL}, the server $c$ aims to train a global model to classify unlabeled samples. We consider a supervised task with unlabeled features $x$ and their ground-truth labels $y$. All devices have the same label $y$, but store different features~$x$. We assume that both devices and the server $c$ store their own datasets imbalanced across labels, i.e., a non-IID global dataset, in which some of the target labels are deficient in samples. For the $k$-th label, $c(k)$ denotes the number of samples in the server $c$. We hereafter refer to these samples as base samples that will be used for data augmentation to make the server's training dataset balanced.

Before augmenting samples, the server $c$ informs its connected devices of its target labels lacking samples, and requests $q^{target}_{c,u_i}$ samples per target label to the device $u_i$. Then, the device uploads XOR encoded samples to the server $c$. At the server $c$, the encoded samples are decoded using XOR operations with the server's base samples, generating a synthetic-but-realistic sample for correcting its imbalanced training dataset. The encoding preserves raw data privacy by mixing target-label samples and dummy-label (non-target-label) samples using the bit-wise XOR operations. Likewise, the decoding preserves data privacy by mixing the encoded sample not with the raw dummy sample but with the server~$c$'s base sample. 

After the decoding, due to the use of the server's base samples, there exist residual noise, as shown in Fig.~\ref{fig:overview}. This is partly intended to preserve privacy, but nonetheless too much noise is obviously harmful for accuracy. To avoid excessive noise, it is important to extract common feature before each encoding or decoding. To this end, up to $p$ samples within the same label are averaged as done in Vanilla Mixup~\cite{Zhang2018}. This sample blending step not only extracts common features, but also preserves more privacy by mixing multiple samples. The aforementioned operations of XorMixFL are elaborated in the following three steps.

\begin{table}[t!]
\footnotesize
\caption{List of Notations
}
\label{tab:param}
\begin{center}
    \resizebox{\columnwidth}{!}{
	\begin{tabular}{c|l}
    \toprule
    Notation & Meaning \\
    \midrule 
    $\mathcal{U}$ & \# of devices \\ 
    $u_i$ & $i$-th device \\ 
    $D_i$ & Dataset of $u_i$\\ 
    $c$ & Server \\ 
    $c(k)$ & \# of base samples in the $k$-th label at $c$\\
    $p$ & Maximum \# of blending samples per label \\
    {$N$} & \# of target labels \\
    {$M$} & \# of dummy labels \\
    {$n$} & \# of samples per each target label \\
    {$m$} & \# of samples in the remaining M dummy labels \\
    {$q^{target}_{c}$} & {Total required \# of samples at $c$} \\
    {$q^{target}_{c, u_i}$} & {\# of samples uploaded to $c$ from $u_i$} \\
    $x^{target}_{(u_i, t)}$ & Raw sample of a target label at $u_i$\\
    $x^{dummy}_{(u_i, t)}$ & Raw sample of a dummy label at $u_i$ \\ 
    $x^{dummy}_{(c, t)}$ & Base sample of a dummy label at $c$ \\ 
    $X^{enc}_{(u_i, c)}$ & Encoded sample at $u_i$, transmitted to $c$ \\  
    $X^{dec}_{(c)}$ & Decoded sample from $X^{enc}_{(u_i, c)}$ at $c$\\
    \bottomrule
	\end{tabular}
	}
\label{tab:param}
\end{center}
    % \vspace{-5mm}
\end{table}

\paragraph{1) Sample Blending.}
Let $x_{(u_i,t)}^{target}$ denote an 1-dimensional vector whose elements are the target labels of device $u_i$ that the server wants to successfully receive. Let $x_{(u_i,t)}^{dummy}$ represent an 1-dimensional vector whose label is not the target one. Per each label, up to $p$ samples are averaged, resulting in $g^{p}(x^{target}_{(u_i, 0)})$ and $g^{p}(x^{dummy}_{(u_i, 0)})$ for target and dummy labels, respectively. Here, $g^{p}(\cdot)$ is given by iteratively applying a sample blending function $g(x_{(u_i, t)})$ by $p$ times, which is defined as
\begin{equation}
    \label{eq:blend}
    g(x_{(u_i, t)}) = \alpha x_{(u_i, t)}+ (1 - \alpha) x_{(u_i, t+1)},
\end{equation} 
where $\alpha \in(0,1)$ denotes the blending ratio of two samples.

% \vspace{-5pt}
\paragraph{2) XOR Encoding.}
For given blended samples $g^{p}(x^{target}_{(u_i, 0)})$ and $g^{p}(x^{dummy}_{(u_i, 0)})$, we apply the bit-wise XOR operation, and obtain an encoded sample $X^{enc}_{(u_i, c)}$, i.e.,
\begin{equation}
    \label{eq:xor}
    X^{enc}_{(u_i, c)} = g^{p}(x^{target}_{(u_i, 0)}) \oplus g^{p}(x^{dummy}_{(u_i, 0)}).
\end{equation} 
The encoded sample $X^{enc}_{(u_i, c)}$ is sent from device $u_i$ to the server $c$.
 
% \vspace{-5pt}
\paragraph{3) XOR Decoding.}

The server~$c$ performs the sample blending operations with its own samples to yield $g^{p}(x^{dummy}_{(c, 0)})$. Given $g^{p}(x^{dummy}_{(c, 0)})$ and the XOR encoded sample $X^{enc}_{(u_i, c)}$ received from $u_i$, the server $c$ applies the bit-wise XOR operation, resulting in the decoded sample~$X^{dec}_{(c, u_i)}$, given as
\begin{equation}
    \label{eq:rexor}
    X^{dec}_{(c, u_i)} = X^{enc}_{(u_i, c)} \oplus g^{p}(x^{dummy}_{(c, 0)}).
\end{equation} 
Note that $X^{dec}_{(c, u_i)}$ is decoded not using the device~$u_i$'s $g^{p}(x^{dummy}_{(u_i, 0)})$ in Eq.~\ref{eq:xor} but using the server's own $g^{p}(x^{dummy}_{(c, 0)})$ in Eq.~\ref{eq:rexor}, thereby preserving the privacy of the raw samples.

To illustrate, as visualized in Fig.~\ref{fig:overview}, consider an example where a server lacking the samples of the target label $5$ out of 10 labels (digits $0, 1, \dots, 8, 9$). To preserve the sample privacy, device 1 selects a dummy label $2$ at random, within which
{$p=2$} 
samples are blended. Likewise, device 1 performs the same operations for the target label $5$. Then, using XOR, device 1 encodes two blended samples $\{2\}$ and $\{5\}$ (see Eq.~\ref{eq:xor}), and sends the encoded sample $\{2\oplus5\}$ to the server. Next, to decode the received sample $\{2\oplus5\}$, the server first blends its own samples in the dummy label~2, creating a dummy sample $\{2'\}$. Then, the server applies XOR to $\{2\oplus5\}$ and $\{2'\}$, yielding the decoded sample $\{5'\}$ (see Eq.~\ref{eq:rexor}). Finally, the server adds $\{5'\}$ into its dataset, and then trains an ML model that is distributed to every device after the training completion. 

Generalizing this to multiple samples, the server requests $q^{target}_{c,u_i}$ encoded samples to its connected device $u_i$. These operations of each device and the server are summarized by Algorithms~\ref{alg:dev} and \ref{alg:edge}, respectively.

\begin{algorithm}[t!]
   \caption{Device's XOR encoding procedure}
   \label{alg:dev}
{\small\begin{algorithmic}
   \STATE {\bfseries Input:} target label, dummy label, $p$, and $q^{target}_{c, u_i}$
   \REPEAT 
   \STATE Randomly select a target-label sample $x^{target}_{(u_i, 0)}$
   \STATE Randomly select a dummy-label sample $x^{dummy}_{(u_i, 0)}$
   \STATE Blend $x^{target}_{(u_i, 0)}$ via Eq.~\ref{eq:blend} up to $p$ times, yielding $g^{p}\!(x^{target}_{(u_i, 0)})$
   \STATE Blend $\!x^{dummy}_{(u_i, 0)}\!$ via Eq.~\ref{eq:blend} up to $p$ times, yielding $g^{p}\!(x^{dummy}_{(u_i, 0)})$
   \STATE XOR the blended samples via Eq.~\ref{eq:xor}, yielding $X^{enc}_{(u_i, c)}$
   
   \STATE Store $X^{enc}_{(u_i, c)}$ in a buffer
   \UNTIL{The buffer size = $q^{target}_{c, u_i}$}
   \STATE Upload the buffer $\{X^{enc}_{(u_i, c)}\}$ to the server~$c$
\end{algorithmic}\normalsize}
\end{algorithm}

%% file: experiment.tex
\section{Experiments}\label{sec:exp}

%\tred{[JH: throughout this section, following the standard definitions, let's consistently use either: (i) (test) accuracy or (ii) per-label accuracy, while removing all our arbitrary definitions such as accuracy of model, average accuracy, etc.]}

%\tred{[JH: Let's consistently use either labels or classes (for other sections, we use the former)]}

%tred{[JH: throughout the paper, consistently use numbers or words; e.g., $3$ target labels vs. three target labels. ]}

In this section, we numerically analyze the performance of the proposed {XorMixFL} scheme in a non-IID MNIST classification task. For the benchmark scheme, we consider standalone ML, Vanilla FL, and MixFL. Unless otherwise specified, by default we consider $U=3$ devices having $n=10$ samples per each target label while storing $m=200$ samples in the remaining $M$ dummy labels.

% . We utilize train and test dataset of MNIST. Standalone ML, and {MixFL} method are used as baseline algorithms.  %\tred{CNN [JH: CNN is a model architecture, while FL and varaitns are training algorithms. To make them consistent with each other, we can say standalone ML, standalong training, etc.]}

\subsection{Accuracy Evaluation under Non-IID Data Distributions} \label{sec:Numaccuracy}
\begin{figure*}[t!]
\centering
\setlength{\tabcolsep}{2pt}
\renewcommand{\arraystretch}{0.2}
\begin{tabular}{ccc}
\includegraphics[page=1, width=0.3\textwidth]{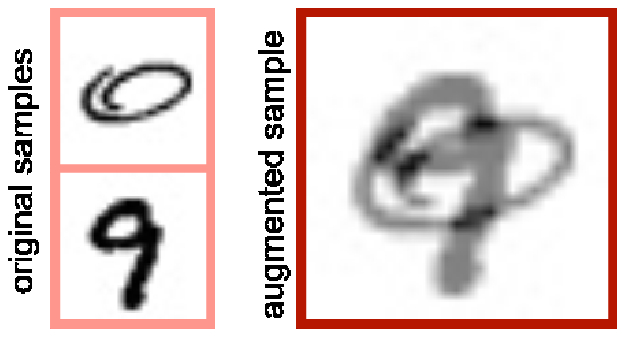} &
\includegraphics[page=1, width=0.25\textwidth]{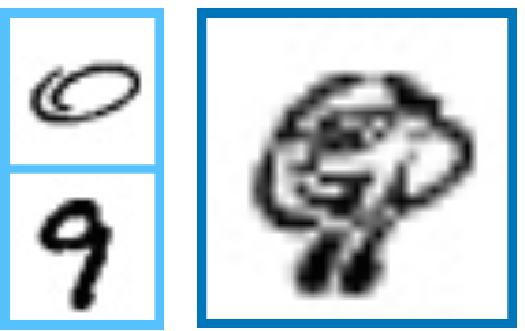} &
\includegraphics[page=1, width=0.40\textwidth]{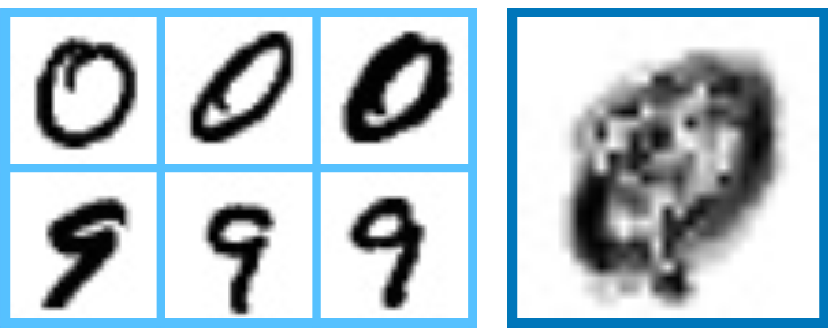} 
\tabularnewline
\tabularnewline
(a) Vanila Mixup. & (b) XorMixup, $p=1$ & (c) XorMixup, $p=3$. \tabularnewline
\end{tabular}
\caption{A visualization of original samples (target label: $0$, dummy label: $9$) and augmented samples using: (a) Vanilla Mixup, (b) XorMixup without sample blending ($p=1$), and (c) XorMixup with sample blending ($p=3$).
% (a) an {augmented} sample and original samples ({target label 0 and dummy label 9}) in case of Vanila Mixup scheme. (b) and (c) represent {augmented} samples and original samples of XorMixup scheme when the number of blending samples $p=1$ and $p=3$, respectively. Here, $p$ means the number of samples which the device/server averages per label. %\tred{[JH: good to explain what $p$ means here]}, respectively. 
}
\label{fig:samples}
\end{figure*}

\begin{figure*}[t!]
\centering
\setlength{\tabcolsep}{2pt}
\renewcommand{\arraystretch}{0.2}
\begin{tabular}{ccc}
\includegraphics[page=1, width=0.33\textwidth]{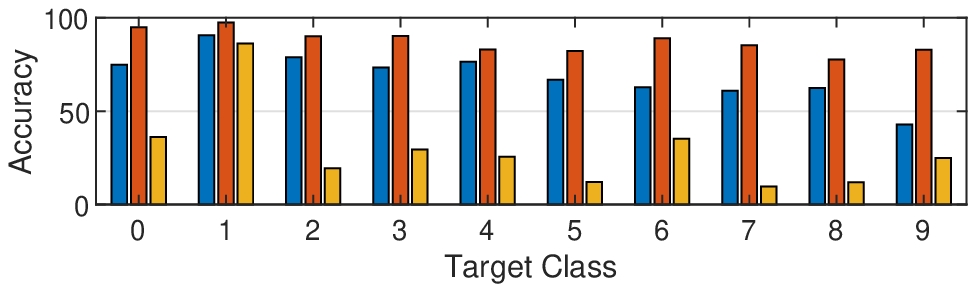} &
\includegraphics[page=1, width=0.33\textwidth]{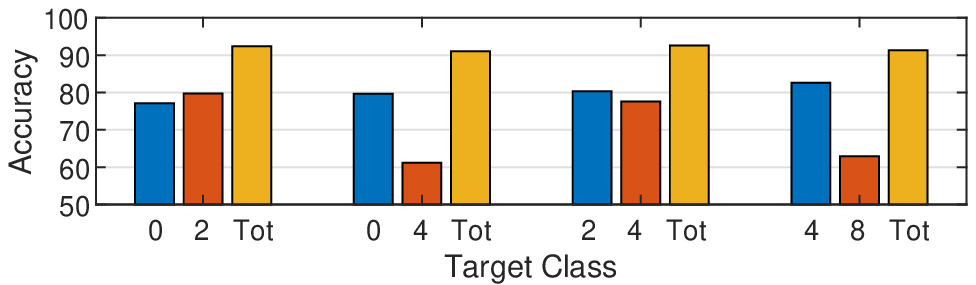} &
\includegraphics[page=1, width=0.33\textwidth]{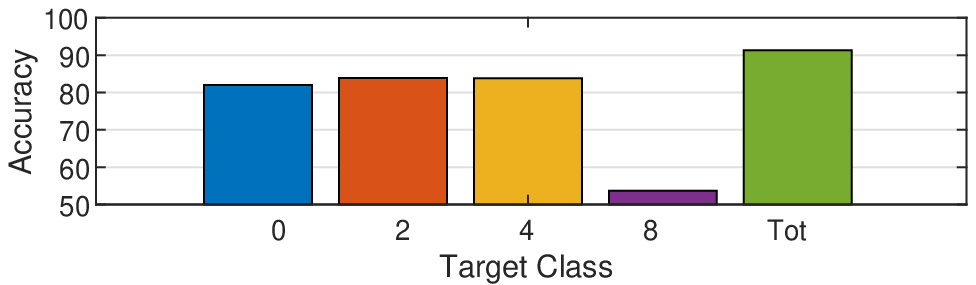} 
\tabularnewline
\tabularnewline
(a) 1 target label ($N=1$)  & (b) 2 target labels ($N=2$) & (c) 4 target labels ($N=4$)
\tabularnewline
\end{tabular}
\caption{ Target-label accuracy evaluation for different {$N\in\{1, 2, 4\}$} $\alpha=0.5$: (a) $N=1$, where the blue, red, yellow bars represent XorMixFL, MixFL, and standalone ML, respectively; (b) $N=2$ (target labels: two of among 0, 2, 4, 8); and (c) $N=4$ (target labels: 0, 2, 4, 8). \textit{Tot} denotes the test accuracy (averaged over all labels).}
\label{fig:xoracc}
% \vspace{-5mm}
\end{figure*}

In this subsection, we investigate the test accuracy and per-label accuracy of XorMixFL and its benchmark schemes, for a different {$N\in\{1, 2, 4\}$}. With $N=1$, Tab.~\ref{tab:def} shows that XorMixFL achieves about 10\% higher test accuracy than the standalone ML regardless of the choice of the target label. Compared to MixFL, XorMixFL shows around 2\% lower test accuracy. This is because the augmented samples generated by XorMixup include more non-trivial noise to preserver more privacy than Vanilla Mixup used in {MixFL}, as visualized in Fig.~\ref{fig:samples}. For each target-label accuracy, Fig.~\ref{fig:xoracc}(a) shows that XorMixFL achievs 65\% target-label accuracy that is 1.87x higher than standalone ML. This highlights the effectiveness of XorMiup in (one-shot) FL.

% First, we consider . First experiment (first to six rows of Tab.~\ref{tab:def}) 
% shows the test accuracy when there is only one target label. 

Next, with $N=2$, Tab.~\ref{tab:def} illustrates that the test accuracy is slightly decreased from the cases with $N=1$, while still exhibiting a similar trend across the proposed and benchmark schemes. To be more specific, comparing Fig.~\ref{fig:xoracc}(a) and (b), target-label accuracy also decreases with $N$, among which the label $0$ is relatively robust against $N$ while the label $4$ is sensitive to $N$. Indeed, the label $4$ accuracy is decreased by around $30$\% compared to the case with $N=1$.

% Table 1
\begin{table*}[t!]
\centering 
\resizebox{\textwidth}{!}{\begin{tabular}{c|c|c|c|c|c|c|c|c|c|c|c|c|c|c|c|c|c|c|c|c}
    \toprule[1.0pt]
    
    \multirow{2}{*}{ \textbf{Method} } & \multicolumn{10}{c|}{ \textbf{1 Target Label}, $N=1$} & \multicolumn{8}{c|}{$N=2$} & \multicolumn{2}{c}{ $N=4$}\\
    \cline{2-21}
     & 0 & 1 & 2 & 3 & 4 & 5 & 6 & 7 & 8 & \multicolumn{1}{c|}{9} & \multicolumn{2}{c|}{(0, 2)} & \multicolumn{2}{c|}{(0, 8)} & \multicolumn{2}{c|}{(2, 4)} & \multicolumn{2}{c|}{(4, 8)} & \multicolumn{2}{c}{($0, \!2, \!4, \!8$)}\\
     \hline
    {XorMixFL} & 94.44\% & 95.83\% & 95.28\% & 93.99\% & 94.72\% & 94.17\% & 94.58\% & 93.43\% & 94.65\% & \multicolumn{1}{c|}{91.54\%} & \multicolumn{2}{c|}{92.36\%} & \multicolumn{2}{c|}{91.03\%} & \multicolumn{2}{c|}{92.61\%} & \multicolumn{2}{c|}{91.28\%} & \multicolumn{2}{c}{88.48\%} \\
    {MixFL} & 96.85\% & 96.72\% & 95.59\% & 96.13\% & 95.82\% & 95.89\% & 95.52\% & 95.57\% & 95.56\% & \multicolumn{1}{c|}{95.34\%} & \multicolumn{2}{c|}{95.52\%} & \multicolumn{2}{c|}{93.98\%} & \multicolumn{2}{c|}{94.87\%} & \multicolumn{2}{c|}{93.76\%}  &
    \multicolumn{2}{c}{91.93\%} \\
    {Vanilla FL} & 83.27\% & 84.83\% & 84.25\% & 83.49\% & 84.82\% & 83.82\% & 85.12\% & 83.31\% & 84.26\% & \multicolumn{1}{c|}{82.94\%} & \multicolumn{2}{c|}{81.72\%} & \multicolumn{2}{c|}{81.23\%} & \multicolumn{2}{c|}{78.75\%} & \multicolumn{2}{c|}{79.23\%}  &
    \multicolumn{2}{c}{77.12\%} \\
      {Standalone} & 89.12\% & 88.62\% & 89.34\% & 91.13\% & 89.83\% & 88.31\% & 89.28\% & 90.11\% & 91.22\% & \multicolumn{1}{c|}{87.43\%} & \multicolumn{2}{c|}{86.41\%} & \multicolumn{2}{c|}{88.16\%} & \multicolumn{2}{c|}{87.19\%} & \multicolumn{2}{c|}{86.58\%}  &
    \multicolumn{2}{c}{84.82\%}\\
    
    \bottomrule[1.0pt]
    \end{tabular}}
    \caption{ Test accuracy evaluation for different {$N\in\{1, 2, 4\}$} ($p = 1$, $\alpha=0.5$).}
\label{tab:def}
\end{table*}

% \tred{[JH: Please make the table fit within 1 row (6 lines). You can simply move the lines at the bottom to the right. As I put the resize command before the tablur, the table will be automatically resized properly.]}

% can see that the test accuracy decreases when the number of deficient label increases. However, per-label accuracy of zero label is relatively more stable than another target label (i.e., 2, 4 and 8). Especially, when label 4 is deficient label with label 0, per-label accuracy of label 4 decreases by about 30\% compared to per-label accuracy of label 4 of Fig.~\ref{fig:xoracc} (a). 

% Except this case, when there are two deficient labels, per-label accuracy of each label is almost similar to that of each label in Fig.~\ref{fig:xoracc}.

%\tred{[JH: is this the third experiment? Please try to maximize the parallel structure. Each paragraph starts in the same way (in terms of the description, notation, figure/table references, etc.)so that one can easily and quickly distinguish the different settings (\#target labels?)]} 
Lastly, with $N=4$, following the same tendency, Tab.~\ref{tab:def} shows that test accuracy further decreases from $N=2$, and the resultant lowest test accuracy is $88.48$\%. Likewise, as observed in Fig.~\ref{fig:xoracc}(c), the target-label accuracy is also the lowest among the cases $N=1$, $2$, and $4$. Specifically, compared to the case with $N=2$ in Fig.~\ref{fig:xoracc}(b), the target-label accuracy is decreased by around $4$\%.

% Third experiment shows that per-label accuracy of {XorMixFL} model when there are four deficient labels. Fig.~\ref{fig:xoracc} (c) . Since there are four deficient labels, the overall performance of the trained model is the lowest among the experiments in Tab.~\ref{tab:def}. Compared to the case of two deficient labels, per-label accuracy of each label is relatively about 4\% lower than that of each label in Fig.~\ref{fig:xoracc} (b). As a result, the test accuracy of four target labels is about 88.48\% as shown in Tab.~\ref{tab:def}.

\subsection{Data Privacy Evaluation}

In this subsection, we study the sample privacy guarantee of the augmented dataset. Following \cite{Jeong:FML19,Oh20:CL}, the sample privacy is measured using the minimum MDS value \cite{Cox2008} between the augmented sample and any raw sample contributing to the augmented sample. A larger MDS value implies higher dissimilarity, preserving more privacy.

Tab.~\ref{tab:mds} shows the MDS values of XorMixFL and MixFL with $N=1$. For the blending ratio $\alpha$, it is trivial that $\alpha$ biased towards either $0$ or $1$ more reveals one of the raw samples, while equal blending ($\alpha=0.5$) minimizes the raw sample leakage. Therefore, for MixFL that shows the highest accuracy in Sec.~\ref{sec:Numaccuracy}, we aim to preserve more privacy by choosing $\alpha=0.5$. By contrast, for XorMixFL, we aim to increase accuracy by choosing $\alpha=0.95$.

\begin{algorithm}[t!]
   \caption{Server's XOR decoding and training procedures}
   \label{alg:edge}
{\small\begin{algorithmic}
   \STATE {\bfseries Input:} $\{X^{enc}_{(u_i, c)}\}$, target label, dummy label, $p$, and  $q^{target}_{c}$
   \STATE Inform connected devices of $q^{target}_{c, u_i}$ and target label
   \REPEAT 
   \STATE Randomly select a dummy-label sample $x^{dummy}_{(c, 0)}$
%   \STATE Find dummy class information from the received data.
%   Select one sample $x^{dummy}_{(c, 0)}$ randomly from the dummy class dataset.
   \STATE Blend $x^{dummy}_{(c, 0)}$ via Eq.~\ref{eq:xor} up to $p$ times, yielding $g^{p}(x^{dummy}_{(c, 0)})$
   \STATE XOR $X^{enc}_{(u_i, c)}$ and $g^{p}(x^{dummy}_{(c, 0)})$ via Eq.~\ref{eq:rexor}, yielding $X^{dec}_{(c, u_i)}$
%   $X^{dec}_{(c, u_i)} = X^{enc}_{(u_i, c)} \oplus x^{dummy}_{(c)}$
   \STATE Store $X^{dec}_{(c, u_i)}$ in the training dataset
   \UNTIL{\# of decoded samples =  $q^{target}_{c}$}
   \STATE Train the server's model
   \STATE Broadcast the server's trained model to all devices
\end{algorithmic} \normalsize}
\end{algorithm}
% \vspace{-1mm}

As shown in Tab.~\ref{tab:mds}, the value of MDS increases with the number $p$ of belnding samples and the number $M$ of dummy labels. Therefore, the lowest mean value of MDS in all cases of {XorMixFL} is $2116.96$, which is even higher than the maximum MDS value $2095.57$ of all cases in {MixFL}. The highest mean MDS value in {XorMixFL} is $3030.43$ when $M=3$ and $p = 5$. The highest value of the MDS of {MixFL} is $2095.57$. The highest MDS value of {XorMixFL} is about 50\% higher than that of {MixFL}. This results show that {XorMixFL} achieves preserves more data privacy than {MixFL}.

\subsection{Impacts of Hyperparameters}
In this subsection, we compare {XorMixFL} and {MixFL} in terms of the test accuracy and sensitivity to key design parameters: the number $M$ of dummy labels, sample blending weight $\alpha$, and the maximum number $p$ of blending samples per label. For all the parameter changes, XorMixFL is more robust against MixFL as elaborated next.

\paragraph{Test Accuracy.}
First, we study the impact of $M$ on test accuracy. With $M=2$ under $p = 1$ and $\alpha = 0.25$, {XorMixFL} achieves the 95.29\% test accuracy. With $M=3$, the test accuracy of {XorMixFL} is 91.60\%. There is only 3.8\% reduction in test accuracy, which is the largest reduction of {XorMixFL} when only the number of dummy labels changes. On the other hand, the test accuracy of {MixFL} is 94.49\% when $M=1$ under $p = 5$ and $\alpha = 0.25$. With $M=3$, the test accuracy of {MixFL} drops to 85.65\%, corresponding to 9.3\% reduction. This is the largest reduction in {MixFL} when only the number of dummy labels changes.

Second, we investigate the impact of $\alpha$ on test accuracy. With $p = 5$ under $M=2$ and $\alpha = 0.5$, the test accuracy of {XorMixFL} is 92.78\%. With $\alpha = 0.25$, the test accuracy increases by around 2\%. This is the largest increase in the test accuracy of {XorMixFL}. With $\alpha = 0.95$, the test accuracy increases by around 5\%. This is the largest increase in the test accuracy of {MixFL}.
% {On the other hand, when $p = 5$ and $M=3$, the test accuracy of {MixFL} is 85.65\%}.

Lastly, we consider the impact of $p$ on  test accuracy. When $p = 5$ under $\alpha = 0.5$ and $M=2$, 92.78\% is the test accuracy of {XorMixFL}. With $p = 1$, the test accuracy increases by about 3\%. However, With $p = 5$ under $\alpha = 0.95$ and $M=1$, the test accuracy of {MixFL} is 90.74\%. When $p = 1$, the test accuracy increases by about 5\%.

%%%%%%%%%%%%%%%%%%%%%%%%%%%%%%%%%%%%%%%%%%%%%%%%%%%%%%%%%%%%%
\begin{table*}[t!]
\footnotesize
\centering
\resizebox{\textwidth}{!}{\begin{tabular}{cc|c|c|c|c|c}
    \toprule[1.0pt]
    %%%%%%%%%%%%%%%%%%%%%%%%%%%
    \multirow{2}{*}{} & & \multicolumn{5}{c}{\textbf{Number of Averaging Samples} ($p$)} \\
    \cline{3-7}
     && $p=1$ & 2 & 3 & 4 & 5\\
    \hline
    %%%%%%%%%%%%%%%%%%%%%%%%%%
    \multirow{3}{*}{{XorMixFL}} & $M=1$  & 
2116.97 / 335.85 &
2335.78 / 545.64 &
2520.88 / 687.77 &
2674.01 / 778.90 &
\textbf{2781.09 / 838.49}\\
    & $M=2$ & 2321.38/ 352.65 & 2409.72 / 643.56 & 2664.12 / 814.13 & 2821.37 / 900.01 & \textbf{2911.42 / 947.06}\\
    & $M=3$ & 2406.62 / 365.61 & 2519.44 / 741.04 & 2800.24 / 895.33 & 2949.51 / 966.04 & \textbf{3030.43 / 1001.48}\\
    \hline
    \multirow{3}{*}{{MixFL}} & $M=1$  &  
    1248.46 / 156.20 &
    1342.34 / 213.23 &
    1382.11 / 311.44 &
    1402.53 / 292.41 &
    \textbf{1496.32 / 349.25}
    \\
    & $M=2$ &  
    1462.75 / 251.78 &
    1523.10 / 320.13 &
    1611.42 / 343.13 &
    1683.45 / 391.39 &
   \textbf{1820.43 / 385.32}
    \\
    & $M=3$ &  
    1521.13 / 225.39 &
    1621.47 / 276.52 &
    1850.32 / 403.72 &
    1999.01 / 545.66 &
    \textbf{2095.57 / 634.79} 
    \\
    \bottomrule[1.0pt]
    \end{tabular}}
    \caption{ MDS comparison (mean / standard deviation) between XorMixFL and MixFL (target label: $9$, $\alpha=0.5$). }
    \label{tab:mds}
\end{table*}
\begin{table*}[t!]
\small
\centering
\resizebox{\textwidth}{!}{\begin{tabular}{c|c|c|ccc|ccc|ccc}
    \toprule[1.0pt]
    %%%%%%%%%%%%%%%%%%%%%%%%%%%
    \multicolumn{3}{c|}{} & \multicolumn{3}{c|}{\textbf{1 dummy label, $M=1$}} & \multicolumn{3}{c|}{\textbf{2 dummy labels}, $M=2$} & \multicolumn{3}{c}{\textbf{3 dummy labels}, $M=3$}\\
    \cline{4-12}
  
    \multicolumn{3}{c|}{} & $\alpha$ = 0.25 & 0.5 & 0.95 & 0.25 & 0.5 & 0.95 & 0.25 & 0.5 & 0.95 \\
    \hline
    %%%%%%%%%%%%%%%%%%%%%%%%%% 
    \multirow{4}{*}{{XorMixFL}} &  \multirow{2}{*}{$p$ = 1} &  Test Acc. & 92.33\%  & 92.54\%  & 92.05\% & \textbf{95.29\%} & 95.10\% & 94.47\%  & 91.60\%& 91.51\%& 90.82\%\\
    % \cline{3-12}
    & & Per-label Acc.  & 52.89\% & 54.48\% & 51.65\% &  \textbf{80.01\%} & 77.61\% & 74.45\% &  44.62\% & 41.98\% & 39.72\%\\
    \cline{2-12}
    \cline{3-12}
    &  \multirow{2}{*}{$p$ = 5} &  Test Acc.  & 91.75\% & 92.98\% & 93.11\% & \textbf{94.37\%} & 92.78\% & 94.19\% & 90.82\%& 90.42\%& 91.60\% \\
    % \cline{3-12}
    & & Per-label Acc. & 46.16\% & 56.55\% & 60.28\% & \textbf{69.59\%} & 56.23\% & 68.95\% & 34.92\%& 33.42\%& 43.68\%\\
    \hline
    \multirow{4}{*}{{MixFL}} &  \multirow{2}{*}{$p$ = 1} &  Test Acc.  & \textbf{94.49\%} & 94.49\% & 95.96\% & 87.85\% & 90.65\% & 91.98\% & 87.44\% & 90.31\% & 91.86\%\\
    % \cline{3-12}
    &                                                & Per-label Acc.   & \textbf{72.82\%} & 72.60\% & 72.73\% & 21.94\% & 43.36\% & 47.72\% & 19.78\% & 40.91\% & 46.46\%\\
    \cline{2-12}
    \cline{3-12}
    &  \multirow{2}{*}{$p$ = 5} &  Test Acc.  & \textbf{94.49\%} & 90.07\% & 90.74\% & 88.64\% & 86.21\% & 90.71\% & 85.65\% & 86.03\% & 90.66\%\\
    % \cline{3-12}
    &                           & Per-label Acc.   & \textbf{42.08\%} & 32.82\% & 35.67\% & 26.83\% & 24.71\% & 34.76\% & 15.42\% & 24.30\% & 34.72\%\\
    \bottomrule[1.0pt]
    \end{tabular}}
    \caption{ The average test accuracy of {XorMixFL} and {MixFL} when the number of dummy label and the value of $\alpha$ for image blending procedure changes  (target label: $9$, dummy labels: $3$ and/or $4$ and/or $5$). }
    \label{tab:acc}
% \tred{dummy labels: $3$ and/or $5$ [JH: up to 2 dummy labels? Then what is $M=3$ in the last column?]}
% \vspace{-5mm}
\end{table*}

\paragraph{Target-Label Accuracy.} In terms of per-label accuracy, we compare the impact of $M$ and $\alpha$. First, when $\alpha = 0.5$ under $p = 1$, and three dummy labels are used, the per-label accuracy of {XorMixFL} is 41.98\%. When $M=2$, the per-label accuracy increase by about 84\%, leading to the per-label accuracy 77.61\%. This is the largest percent change in {XorMixFL} when only the number of dummy labels changes. On the other hand, when $\alpha = 0.25$ under $p = 1$ and $M=3$, the per-label accuracy of {MixFL} is 19.78\%. When $M=1$, the per-label accuracy increase by about 268\%, resulting in the per-label accuracy 72.82\%. This is the largest percent change in {MixFL} when only the number of dummy labels changes.

Second, we study the impact of the blending weight $\alpha$ on the per-label accuracy. When $p = 5$ under $\alpha = 0.25$ and $M=1$, the per-label accuracy of {XorMixFL} is 46.16\%. When $\alpha = 0.95$, the per-label accuracy increase by about 30\%, resulting in the per-label accuracy is 60.28\%. This is the largest increase in the per-label accuracy of {XorMixFL} when only $\alpha$ changes. On the other hand, when $p = 1$ under $\alpha = 0.25$ and $M=3$, the per-label accuracy of {MixFL} is 19.78\%. When $\alpha = 0.95$, the per-label accuracy increases by about 134\%, leading to the per-label accuracy 46.46\%.

Lastly, we investigate the impact of $p$ on the per-label accuracy. With $p = 5$ under $\alpha = 0.5$ and $M=2$, the per-label accuracy of {XorMixFL} is 56.23\%. The per-label accuracy increases by about 38\% when $p = 1$, yielding the per-label accuracy 77.61\% of {XorMixFL}. However, with $p = 5$ under $\alpha = 0.25$ and $M = 1$, the per-label accuracy of {MixFL} is 42.08\%. With $p = 1$, the per-label accuracy increases by about 73\%, achieving 72.82\% per-label accuracy of {MixFL}. 

% As a result, {XorMixFL} is more robust than {MixFL} in terms of aforementioned three variables. Therefore, {XorMixFL} can obtain more stable performance while increasing the degree of privacy. 

%% file: conclusion.tex
\section{Conclusion}\label{sec:con}
In this work, we proposed a novel privacy-preserving one-shot FL framework, XorMixFL, which allows devices to locally augment insufficient samples to correct the non-IID data distributions while hiding the details of the original samples. Numerical simulations validate the effectiveness of XorMixFL in terms of accuracy and privacy guarantees, while discovering the impacts of key design parameters such as the numbers of blending samples and dummy labels on accuracy and privacy guarantees. Exploiting the core idea, XorMixup data augmentation, it could be interesting to extend our one-shot FL framework to the standard multi-shot FL applications for future study.